\documentclass[preprint,12pt]{elsarticle}

\usepackage{amssymb}
\usepackage{csquotes}
\usepackage{float}
\usepackage{multirow}
\usepackage{booktabs} 
\usepackage[table,xcdraw]{xcolor}
\usepackage{amsmath}

\journal{Smart Agricultural Technology }

\begin{document}

\begin{frontmatter}



\title{Lightweight Model for Poultry Disease Detection from Fecal Images Using
Multi-Color Space Feature Optimization and Machine Learning}

\author[inst1]{A K M Shoriful Islam\corref{cor1}} 
\cortext[cor1]{Corresponding author}
\ead{shorifulislamsimul@gmail.com}
\author[inst1]{Md. Rakib Hassan} 
\author[inst1]{Macbah Uddin} 
\author[inst2]{Md. Shahidur Rahman} 

\affiliation[inst1]{organization={Department of Computer Science and Mathematics, Faculty of Agricultural Engineering \& Technology \\ Bangladesh Agricultural University},
            city={Mymensingh},
            postcode={2202}, 
            country={Bangladesh}}

\affiliation[inst2]{organization={Department of Poultry  Science, Faculty of Animal Husbandry\\ Bangladesh Agricultural University},
            city={Mymensingh},
            postcode={2202},
            country={Bangladesh}}

\begin{abstract}
Poultry farming is a vital component of the global food supply chain, yet it remains highly vulnerable to infectious diseases such as coccidiosis, salmonellosis, and Newcastle disease. This study proposes a lightweight machine learning-based approach to detect these diseases by analyzing poultry fecal images. We utilize multi-color space feature extraction (RGB, HSV, LAB) and explore a wide range of color, texture, and shape-based descriptors, including color histograms, local binary patterns (LBP), wavelet transforms, and edge detectors. Through a systematic ablation study and dimensionality reduction using PCA and XGBoost feature selection, we identify a compact global feature set that balances accuracy and computational efficiency. An artificial neural network (ANN) classifier trained on these features achieved 95.85\% accuracy while requiring no GPU and only 638 seconds of execution time in Google Colab. Compared to deep learning models such as Xception and MobileNetV3, our proposed model offers comparable accuracy with drastically lower resource usage. This work demonstrates a cost-effective, interpretable, and scalable alternative to deep learning for real-time poultry disease detection in low-resource agricultural settings.
\end{abstract}


\begin{highlights}
\item Lightweight model for poultry disease detection using fecal images.
\item Multi-color space features (RGB, HSV, LAB) enhance classification accuracy.
\item Global feature set selected via ablation and feature selection techniques.
\item ANN classifier achieved 95.85\% accuracy without GPU support.
\item Suitable for low-resource deployment in real-time poultry farms.
\end{highlights}

\begin{keyword}
Poultry disease detection, machine learning, fecal image classification, color space analysis, lightweight models.
\end{keyword}

\end{frontmatter}

\section{Introduction}
\label{chap:intro}
Poultry is one of the most efficient and highly cultivated programs to sustain the regular effort of protein demand. The population is rapidly increasing worldwide, and we will face an unbalanced protein supply \cite{daghir2021poultry}. Poultry could be the most efficient solution if its vulnerability could be eradicated, which happens because of its disease tendency.

Poultry disease varies depending on weather, season and age. \cite{islam2003study} found that at 1-7 days, 13.68\% of chickens are affected with \textit{Aspergillosis}; for 8-35 days old, 11.32\% of chickens are infected with \textit{Bursal Disease} (IBD), and at 35-60 days old 1.11\% of poultry are infected with \textit{Newcastle} disease. In the season consideration, the rainy season is the most prevalent season of poultry disease (56.3\%), followed by summer (28.11\%), and the least in the winter season (15.53\%).
Another study in Pakistan  \cite{anjum1990weather} found that the most common disease in poultry is \textit{Newcastle} disease (37.6\%), hydropericardium syndrome(17.3\%), feed poisoning (12.8\%) coccidiosis (15\%), Colisepticemia (8.9\%), salmonellosis (8.4\%), infectious bronchitis (3.1\%), infectious coryza  (1.3\%), fowl cholera  (0.9\%), mycoplasmosis (5.8\%), avian leukosis (2.7\%), avipoxvirus (2.2\%), contagious laryngotracheitis (0.95\%), and inclusion body hepatitis (1.3\%). Newcastle disease was more prevalent in layers (51.6\%) than in broilers (27.8\%), while hydropericardium syndrome and feed intoxication were more common in broilers than in layers. Diseases such as Newcastle disease, salmonellosis, coccidiosis and avipoxvirus infection were more prevalent in spring than in other seasons, and feed intoxication was commonest in summer.
\cite{uddin2010prevalence} found that the most common diseases in Bangladesh are Gumboro Disease, Newcastle Disease, Colibacillosis and Salmonellosis. 

\section{Related Work}
Early detection of poultry diseases is vital for flock health and industry sustainability. Traditional methods relying on manual inspection and lab testing are often time-consuming and costly, prompting the adoption of automated solutions using artificial intelligence (AI).

Sound-based detection methods have shown promise. Banakar et al. \cite{banakar2016intelligent} used the Dempster-Shafer theory and SVM to classify sound features, achieving up to 91.15\% accuracy. Carpentier et al. \cite{carpentier2019development} detected sneezing sounds with 88.4\% precision using Linear Discriminant Analysis. Cuan et al. introduced CSCNN and BiLSTM models for detecting avian influenza and Newcastle disease, achieving accuracies up to 98.5\% \cite{cuan2020detection, cuan2022automatic}. Hassan et al. \cite{hassan2024optimizing} enhanced classification using a “Burn Layer” in a CNN model, reaching 98.55\% accuracy.

Image-based methods using poultry droppings have been explored extensively. Degu et al. \cite{degu2023smartphone} combined YOLO-V3 and ResNet50 to classify diseases from fecal images with 98.7\% accuracy. Suthagar et al. \cite{suthagar2023faecal} achieved 97\% accuracy using DenseNet. Wang et al. \cite{wang2019recognition} applied CNNs to detect digestive issues based on dropping morphology, reporting 93.3\% mean average precision. Machuve et al. \cite{machuve2022poultry} and Qin et al. \cite{qin2024deep} further enhanced detection using deep learning and object tracking, attaining up to 99.45\% accuracy with fine-tuning.

Posture recognition is another promising avenue. Zhuang et al. \cite{zhuang2019detection} used InceptionV3 and feature fusion to detect sick broilers with 99.7\% mAP. Okinda et al. \cite{okinda2019machine} used posture and walking speed, achieving over 97\% accuracy with SVM. Fang et al. \cite{fang2021pose} applied pose estimation to classify chicken behaviors, although illness detection was not addressed. Ahmed et al. \cite{ahmed2021approach} used IoT-based accelerometer data and TabNet for disease classification, reporting 97\% accuracy.

These studies highlight the potential of multimodal AI methods in disease detection. However, challenges remain in generalization, field deployment, and detecting multiple diseases across age groups and environmental conditions. More scalable, accurate, and disease-specific models are needed for real-world poultry health monitoring.

\section{Working Procedure}
We have a dataset of RGB poultry-dropping images having four classes. Among the
datasets, three classes are disease-affected, and another class contains healthy photos.
First, we converted RGB images into HSV and LAB modes. Secondly, including RGB
converted HSV and LAB mode decomposed into ten different channels (Gray, Red,
Green, Blue, Hue, Saturation, Violet, L, A, B). Thirdly, essential features are extracted from those channels via ablation study. Fourthly, we applied several machine learning algorithms
to train our classifier model. Finally, a test set of images was used to explore the model's
performance. 

\begin{figure}[H]
    \centering
    \includegraphics[width=1\linewidth]{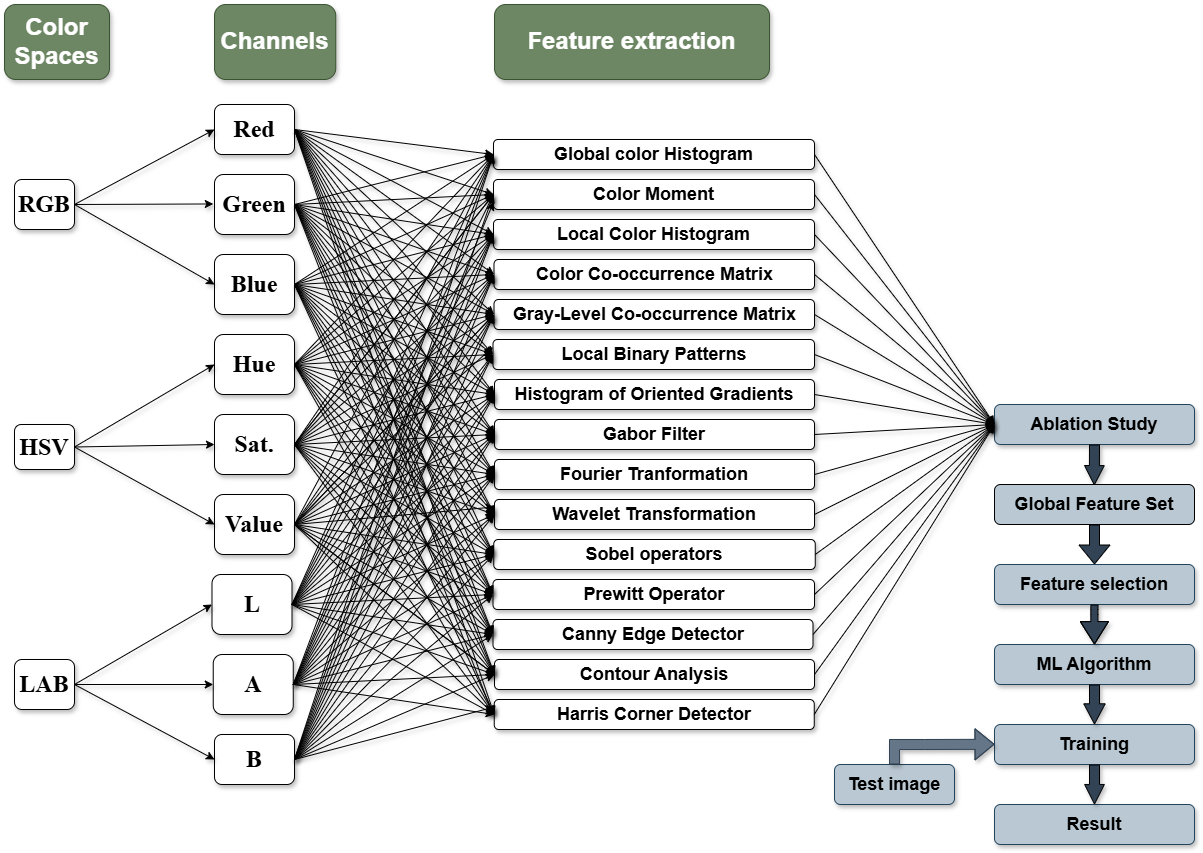}
    \caption{Proposed Model}
    \label{fig:my_label1}
\end{figure}

The process is accomplished using Python, as described in the Figure \ref{fig:my_label1}.

\subsection{Dataset Description}
This research collected the dataset from ZENODO REPOSITORY \cite{machuve2021machine}. We have found that there are two types of datasets:
($i$) Farm-labeled fecal images,
($ii$) Laboratory-labeled fecal images. Each contains four classes: \enquote{Salmo}, \enquote{ncd}, \enquote{healthy} and \enquote{cocci}. We combined these two datasets into a single dataset that gives 8067 images. \begin{table}[H]
    \centering
    \begin{tabular}{|l|l|}
        \hline
        \multicolumn{2}{|c|}{Dataset}                                                    \\ \hline
        Class        & Total number of images                                               \\ \hline
        Coccidiosis  & 2476                                                                  \\ \hline
        Healthy      & 2404                                                                  \\ \hline
        Newcastle    & 562                                                                   \\ \hline
        Salmonellsis & 2625                                                                  \\ \hline
    \end{tabular}
    \caption{Number of images for different classes.}
    \label{tab:data_set_image_count}
\end{table}
\subsection{Image Mode Conversion}
We used \texttt{cv2.COLOR\_BGR2HSV} and \texttt{cv2.COLOR\_BGR2LAB} functions to convert RGB images into HSV and CIE L{*}A{*}B colour spaces, respectively. Different colour spaces can significantly impact poultry droppings classification. While RGB is widely used for image display, it struggles to separate luminance from chromatic information~\cite{foley1996computer}. HSV, better aligned with human colour perception, often improves segmentation and colour-based analysis~\cite{hema2019interactive}, though its effectiveness may vary depending on the dataset. CIE L{*}A{*}B, being perceptually uniform, excels at capturing subtle colour differences and separating luminance from colour.

\begin{figure}[H]
    \centering
    \includegraphics[width=0.8\linewidth]{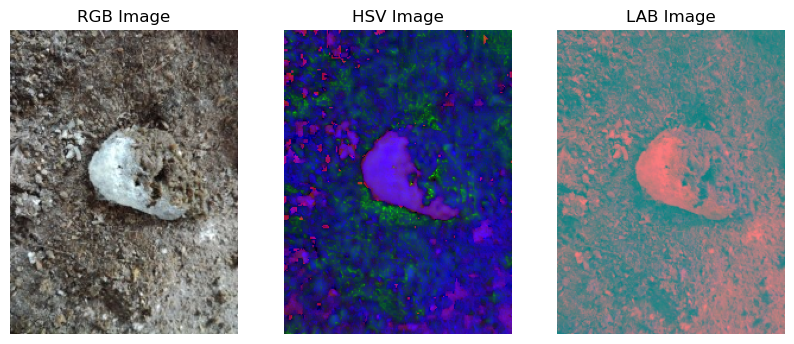}
    \caption{Color space conversion from RGB image}
    \label{fig:my_label2}
\end{figure}

\subsection{Feature Extraction}
Feature extraction is one of the most important tasks of our work. For poultry fecal image classification, our main obstacles are noise, irregular shape, non-uniform patterns, and missing or Inconsistent features of images. We focused on colour distribution, texture,  shape or structure analysis to select the best features for the proper classification (Table \ref{tab:feature_summary})

\begin{table}[H]
\begin{tabular}{|c|l|}
\hline
Analysis                                              & \multicolumn{1}{c|}{Feature extraction}                            \\ \hline
\multirow{5}{*}{Color Analysis}                       & CH - Color Histogram                                               \\ \cline{2-2} 
                                                      & CM1 - Color Moments (Mean, Standard Deviation)                     \\ \cline{2-2} 
                                                      & CM2 - Color Moments (CM1, Skewness, Kurtosis) \\ \cline{2-2} 
                                                      & LCH- Local Color Histogram                                         \\ \cline{2-2} 
                                                      & CCM - Co-occurrence Color Matrix                                   \\ \hline
\multirow{6}{*}{Texture Analysis} & GLCM-Gray level co-occurrence Matrix \\ \cline{2-2} 
                                                      & LBP- LocalBinary Pattern                                           \\ \cline{2-2} 
                                                      & HOG - Histogram Oriented Gradient                                  \\ \cline{2-2} 
                                                      & GF-Gabor Filter                                                    \\ \cline{2-2} 
                                                      & FT-Fourier Transformation                                          \\ \cline{2-2} 
                                                      & WT-Wavelet Transformation                                          \\ \hline
\multicolumn{1}{|l|}{\multirow{5}{*}{Shape Analysis}} & Sobel operator                                                     \\ \cline{2-2} 
\multicolumn{1}{|l|}{}                                & Prewitt operator                                                   \\ \cline{2-2} 
\multicolumn{1}{|l|}{}                                & Canny edge detector                                                \\ \cline{2-2} 
\multicolumn{1}{|l|}{}                                & Contour Analysis                                                   \\ \cline{2-2} 
\multicolumn{1}{|l|}{}                                & Harris Corner detector                                             \\ \hline
\end{tabular}
\label{tab:feature_summary}
\caption{List of feature extraction methods used for different image analysis tasks.}
\end{table}

\subsection{Classification Algorithms}
We have applied several machine learning algorithms for classification and used ANN to
identify relatively essential features.
After selecting the important features, we evaluated four machine learning classifiers on the selected feature subset: (1) An Artificial Neural Network (ANN) with 256-128-4 architecture employing ReLU activation (hidden layers) and softmax output, optimized via Adam (learning rate=0.001) with 50\% dropout regularization to mitigate overfitting; (2) Decision Trees utilizing Gini impurity for splitting with max depth=15 and minimum samples leaf=5; (3) Random Forest comprising 100 trees with bootstrap sampling and $\sqrt{n}$ feature randomization per split, leveraging ensemble majority voting; and (4) Support Vector Machines (SVM) with RBF kernel ($\gamma=0.1$, $C=1.0$) employing one-vs-rest multi-class strategy, preceded by feature standardization. All implementations used scikit-learn (v1.0.2) except the ANN (TensorFlow 2.4), with hyperparameters tuned via 5-fold cross-validation to optimize generalization performance across all four poultry disease classes.

\subsection{Hardware description}
We applied several color, texture, and shape analysis procedures to find the vital feature extraction methods. We used a local computer for this analysis, which has the configuration of an Intel i7 processor with 4 cores and 8 threads with 4.2GHz. The RAM specification is a 16GB DDR4 2330 MHZ bus.

After finding the important feature extraction method, we built our ML model. To compare it with the existing deep learning model, we ran all the deep learning models and the proposed model in Google Colab Pro with a T4 high RAM GPU.

\section{Experimental Results}
In this section, we present and discuss the results of our experiments on detecting poultry diseases using machine learning. The main goal of these experiments is to understand how well our model can identify different diseases based on features extracted from fecal images.

First, we identified the best features from different color spaces—RGB, HSV, and LAB—to see how each affects the model’s accuracy. Next, we combine color, texture, and shape features to improve detection accuracy. Finally, we compare our model's results with those of other deep learning models to show how effective our approach is in terms of both accuracy and efficiency.

The results from these experiments help us understand our model's strengths and limitations and give us ideas for future improvements. Each section provides a detailed look at the findings and their definitions for poultry disease detection.

\subsection{Feature Selection From Color Spaces}
The study employed a systematic pipeline for feature optimization and classification. Multi-channel features (color, texture, shape) were extracted from poultry fecal images across four disease classes (cocci, healthy, ncd, salmo), normalized via Min-Max scaling, and dimensionally reduced to 300 principal components using PCA. An 80-20 train-test split was applied with one-hot encoded labels. The classification employed a 3-layer ANN (128-64-4 neurons, ReLU/softmax activation) with dropout regularization (p=0.5), trained for 100 epochs (batch size=32) using Adam optimization and early stopping (patience=5). Model performance was evaluated through standard metrics (accuracy, precision, recall, F1) and confusion matrix analysis on the test set.

\begin{table}[H]
\centering
\resizebox{\textwidth}{!}{%
\begin{tabular}{|p{3cm}|p{4cm}|c|c|c|c|}
\hline
\multicolumn{2}{|c|}{\multirow{2}{*}{\textbf{Feature Testing}}} &
  \multicolumn{4}{c|}{\textbf{Classification Accuracy using ANN (\%)}} \\
\multicolumn{2}{|c|}{} &
  \textbf{RGB} & \textbf{HSV} & \textbf{LAB} & \textbf{RGB+HSV+LAB} \\ \hline

\multirow{7}{*}{\textbf{Color analysis}} & CH & 84.32 & 90.71 & 90.15 & 91.33 \\ \cline{2-6}
& CM1 & 79 & 80.73 & 82.71 & \\ \cline{2-6}
& CM & 79.06 & 84.45 & 84.08 & 91.33 \\ \cline{2-6}
& LCH & 81.16 & 91.82 & 90.02 & 90.83 \\ \cline{2-6}
& CCM & 80.92 & 91.08 & 90.15 & 88.66 \\ \cline{2-6}
& CH+LCH & 82.96 & 91.26 & 90.46 & 91.82 \\ \cline{2-6}
& CH+CM2+LCH+CCM & 84.01 & 92.32 & 91.39 & 91.02 \\ \hline

\multirow{10}{*}{\shortstack{\textbf{Texture}\\\textbf{analysis}}} & GLCM & 58.33 & 75 & 66.66 & \\ \cline{2-6}
& LBP & 75.03 & 91.7 & 89.47 & 92.62 \\ \cline{2-6}
& HOG & 72.12 & 86.06 & 81.04 & 87.67 \\ \cline{2-6}
& GF & 56.13 & 58.67 & 58.67 & \\ \cline{2-6}
& FT & 78.44 & 84.51 & 90.27 & 85.94 \\ \cline{2-6}
& WT & 80.55 & 82.53 & 84.51 & 85.81 \\ \cline{2-6}
& LBP+FT & 71.44 & 81.78 & 87.79 & 74.35 \\ \cline{2-6}
& LBP+WT & 78.93 & 85.75 & 84.94 & 84.94 \\ \cline{2-6}
& FT+WT & 81.78 & 87.11 & 90.27 & 90.46 \\ \cline{2-6}
& LBP+FT+WT & 80.98 & 89.59 & 91.2 & 85.87 \\ \hline

\multirow{6}{*}{\shortstack{\textbf{Shape}\\\textbf{analysis}}} & Sobel operator & 73.3 & 85.19 & 85.56 & 85.81 \\ \cline{2-6}
& Prewitt operator & 72.43 & 84.39 & 84.45 & 86.49 \\ \cline{2-6}
& Canny edge detector & 66.79 & 80.42 & 72.61 & 78.44 \\ \cline{2-6}
& Contour Analysis & 33.21 & 35.32 & 33.15 & 35.56 \\ \cline{2-6}
& Harris Corner & 41.57 & 66.54 & 65.18 & 70.2 \\ \cline{2-6}
& Sobel + Prewitt + Canny & 71.62 & 70.2 & 70.14 & 70.26 \\ \hline
\end{tabular}%
}
\caption{Classification accuracy on different color spaces and features}
\label{tab:color_space_accuracy}
\end{table}

\begin{figure}[H]
     \centering
     \includegraphics[width=\linewidth]{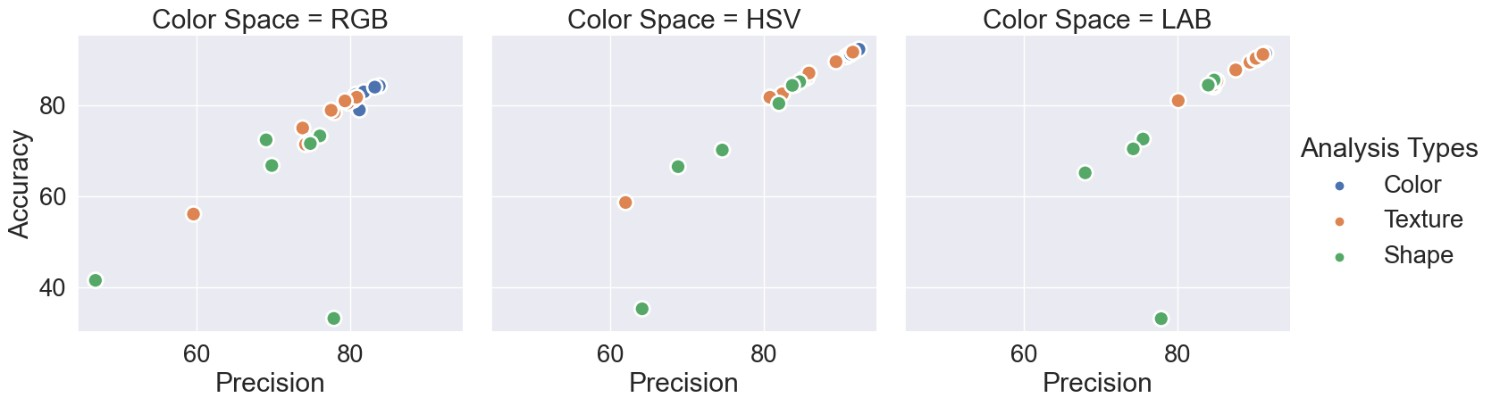}
     \caption{Accuracy comparison on color channels }
     \label{fig:Accuracy comparison on color channels}
     \end{figure}

Table \ref{tab:color_space_accuracy} demonstrates the classification accuracy from different feature settings which shows HSV and LAB channels perform better than the RGB channels visualized in Figure \ref{fig:Accuracy comparison on color channels}. First, the extracted features are classified with ANN, then the accuracy and precision are plotted with the help of the Seaborn Python library. RGB has a maximum accuracy of 84.32\%. However, HSV and LAB have a maximum accuracy of 92.32\%  and 91.39\%, respectively. 

On the other hand, texture analysis has a better accuracy of 93.62\% than color and shape analysis of 92.32\% and 86.49\%, respectively and the scenario of the color, texture and shape analysis are visualized in figure \ref{fig:Accuracy comparison color, shape and texture analysis}.
 \begin{figure}[H]
     \centering
     \includegraphics[width=\linewidth]{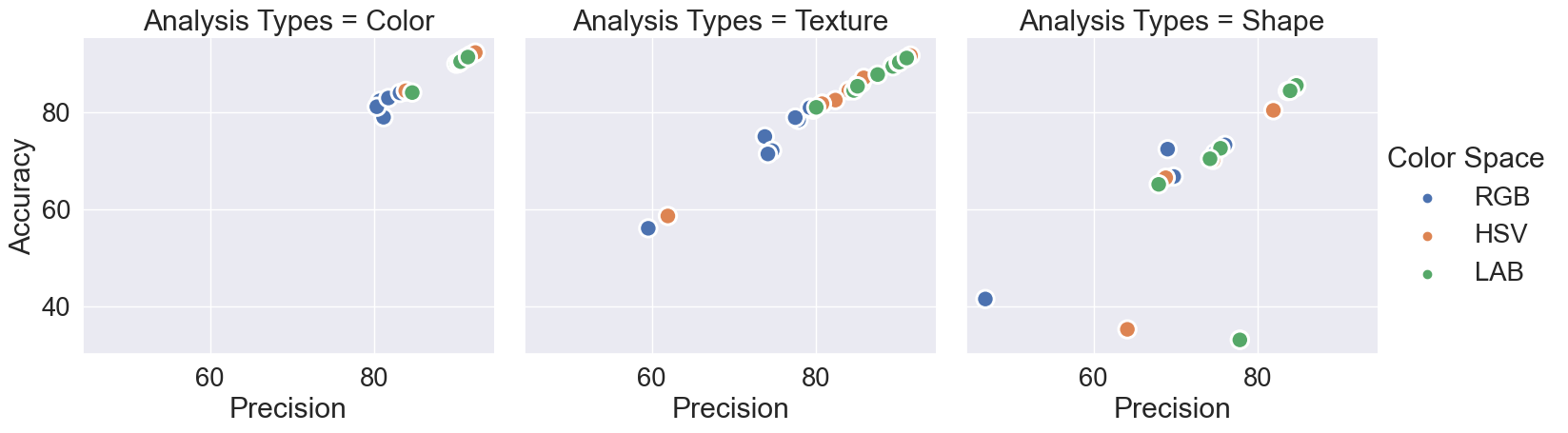}
     \caption{Accuracy comparison color, shape and texture analysis }
     \label{fig:Accuracy comparison color, shape and texture analysis}
     \end{figure}
Figures \ref{fig:Accuracy comparison on color channels} and \ref{fig:Accuracy comparison color, shape and texture analysis} are plotted with all accuracy data for the set of characteristics found in table \ref{tab:color_space_accuracy}.

\subsection{Combination of Features From Different Color Spaces}
We build different settings from the table \ref{tab:color_space_accuracy} by choosing high-accuracy feature selection methods and combining them in various ways. We conducted this process in three segmentations, i.e., $(i)$ color analysis: a combination of different color features, $(ii)$ texture analysis, the combination of various textures, $(iii)$ shape analysis, the combination of varying shape features from multiple color spaces.


\subsection{Color Feature Optimization and Exclusion Justification }
From the table \ref{tab:color_space_accuracy}, We can see that HSV and LAB color space has higher classification accuracy than the RGB. The maximum accuracy is 92.32\% achieved in HSV CH+CM2+LCH+CCM combination, and the minimum accuracy was achieved at 79\% in RGB CM1. For further analysis, we conducted ablation study to find out the best dominant features in color analysis. 
\begin{table}[H]
\centering
\begin{tabular}{|l|c|c|p{5cm}|}
\hline
\rowcolor{gray!15}
\textbf{Removed Feature} & \textbf{Accuracy} & \textbf{Drop} & \textbf{Interpretation} \\
\hline
HSV-CM & 0.9300 & -0.0093 & Removing this improved accuracy the most. It hurts performance and may be redundant or noisy. \\
\hline
LAB-CH & 0.9275 & -0.0068 &  Also increased accuracy. Not helpful in this model. Possibly redundant with other histogram features. \\
\hline
RGB-LCH & 0.9269 & -0.0062 & Removal improves accuracy; feature may be unnecessary or conflicting.\\
\hline
RGB-CH & 0.9250 & -0.0043 & Removing helped slightly. Indicates RGB histograms may not align well with other features. \\
\hline
HSV-CH & 0.9232 & -0.0025 &  Marginal improvement; may not be essential. \\
\hline
LAB-LCH & 0.9226 & -0.0019 &  Minimal improvement; could be removed if optimization is needed. \\
\hline
HSV-CCM & 0.9213 & -0.0006 &  Slight increase; not impactful either way. \\
\hline
RGB-CM & 0.9213 & -0.0006 &  Same as above.  \\
\hline
LAB-CM & 0.9207 & 0.0000 & No change; neutral, possibly low-impact feature. \\
\hline
HSV-LCH & 0.9195 & 0.0012 & Slight decrease in performance. \\
\hline
RGB-CCM & 0.9170 & 0.0037 &  Removing causes noticeable accuracy loss; important feature.\\
\hline
\end{tabular}
\caption{Ablation study on color features}
\label{tab:ablation_color}
\end{table}

The table \ref{tab:ablation_color} demostrates LAB-CM, HSV-LCH and RGB-CCM is the color features that has positive dominance . So we excluded all the other color features and keep this three feature and will apply classification algorithm on this three feature comabination set. We named this three important feature set as Global Feature Set.

\begin{table}[H]
\centering
\resizebox{\textwidth}{!}{%
\begin{tabular}{|l|l|l|l|l|l|}
\hline
Setting Name & Feature Extraction & Accuracy & Precision & Recall & F1 \\ \hline
\multirow{3}{*}{Color-Set} & RGB-CCM & \multirow{3}{*}{92.44} & \multirow{3}{*}{92.41} & \multirow{3}{*}{92.44} & \multirow{3}{*}{92.42} \\ \cline{2-2}
             & HSV-LCH            &          &            &        &    \\ \cline{2-2}
             & LAB-CM             &          &            &        &    \\ \hline
\end{tabular}
}
\caption{Performance of different feature extraction settings}
\label{tab:feature_extraction_results}
\end{table}

\subsection{Texture Feature Optimization and Exclusion Justification}
 Like color analysis, HSV and LAB color spaces have better texture analysis accuracy than RGB. Considering all the texture feature except HOG, GLCM and GF as they require high RAM. However, we performed ablation study on LBP, FT and WT on the three channel RGB, HSV and LAB.\\
\begin{table}[H]
\centering

\begin{tabular}{|l|c|c|p{5cm}|}
\hline
\rowcolor{gray!15}
\textbf{Removed Feature} & \textbf{Accuracy} & \textbf{Drop} & \textbf{Interpretation} \\
\hline
None (Full Model) & 0.8792 & 0.0000 & Baseline accuracy. \\
\hline
RGB-LBP & 0.8841 & -0.0050 & Removing improved accuracy. Feature is noisy or redundant. \\
\hline
{HSV-LBP} & 0.8767 & 0.0025 & Mild performance drop — useful, low-cost feature. \\
\hline
{LAB-LBP} & 0.8705 & 0.0087 & Notable accuracy drop — important LBP feature. \\
\hline
RGB-FT & 0.8724 & 0.0068 & Moderate drop; FT is expensive to compute. \\
\hline
HSV-FT & 0.8643 & 0.0149 & High drop, but very memory-intensive feature. \\
\hline
LAB-FT & 0.8606 & 0.0186 & Largest drop, but high RAM usage. \\
\hline
RGB-WT & 0.8804 & -0.0012 & Slight gain in accuracy — safe to exclude. \\
\hline
HSV-WT & 0.8773 & 0.0019 & Small performance drop — optional. \\
\hline
LAB-WT & 0.8730 & 0.0062 & Moderate drop — optional but costly. \\
\hline
\end{tabular}
\label{tab:textablation}
\caption{Ablation study on texture Features}
\end{table}
The texture analysis initially included features such as LBP, FT, and WT across RGB, HSV, and LAB color spaces (Table \ref{tab:textablation}). While these texture features helped enhance classification, further ablation analysis and resource profiling revealed key limitations in certain feature groups.

Specifically, FFT and WP features across all channels incurred significant RAM usage and computational overhead, often requiring more than 16GB of memory to process efficiently. In contrast, LBP features were computationally lightweight and offered meaningful performance contributions.

The ablation study shows that removing LAB\_LBP caused a performance drop of 0.87\%, and removing HSV\_LBP caused a smaller drop of 0.25\%, confirming their importance in the model. Meanwhile, excluding FFT and WP features caused performance drops between 0.62\% and 1.86\%, but with a disproportionately high memory cost. Therefore, we chose to keep only HSV\_LBP and LAB\_LBP in the final feature set, ensuring a balanced trade-off between accuracy and computational efficiency.
\begin{table}[H]
\centering
\begin{tabular}{|l|l|l|l|l|l|}
\hline
Setting Name                 & Feature Extraction & Accuracy               & Precision              & Recall                 & F1                     \\ \hline
\multirow{2}{*}{Texture set} & HSV-LBP            & \multirow{2}{*}{93.49} & \multirow{2}{*}{93.45} & \multirow{2}{*}{93.49} & \multirow{2}{*}{92.42} \\ \cline{2-2}
                             & LAB-LBP            &                        &                        &                        &                        \\ \hline
\end{tabular}
\caption{Performance of different feature extraction settings on texture analysis}
\end{table}

\subsection{Feature Optimization and Exclusion Justification in Shape Analysis}
The shape feature analysis, including Sobel, Prewitt, and Canny operators across RGB, HSV, and LAB color spaces, consistently showed lower classification accuracy compared to color and texture features. 

In addition to the performance concern, shape features also impose a considerable computational burden. When included, the model exceeded the available 16GB of RAM during training, leading to out-of-memory errors. This high memory demand stems from large intermediate data generated by edge and contour operations across multiple channels.

Considering both the low contribution to classification performance and the high computational cost, we decided to exclude all shape features from the final model. This not only reduced memory usage significantly but also simplified the feature extraction pipeline without compromising accuracy.
\subsection{Color + Texture Feature Set Optimization}

To fine-tune the final feature set, we conducted an ablation study combining dominant color and texture features. The full model achieved an accuracy of 93.87\%. The removal of LAB-LBP resulted in the most significant accuracy drop of 1.12\%, indicating it is the most dominant and indispensable feature in the current setup. Other features such as HSV-LBP and LAB-Moments showed minor drops (0.06\% and 0.25\% respectively), suggesting they offer some contribution to performance.

In contrast, the removal of HSV-LCH unexpectedly increased accuracy by 0.56\%, implying that it may introduce noise or redundancy. Similarly, excluding RGB-CCM slightly improved performance, supporting the decision to drop it from the final configuration.

Based on this, we retain LAB-LBP, LAB-CM, and HSV-LBP in the final model, while excluding HSV-LCH and RGB-CCM to enhance accuracy and efficiency.

\begin{table}[H]
\centering

\begin{tabular}{|l|c|c|p{5cm}|}
\hline
\rowcolor{gray!15}
\textbf{Removed Feature} & \textbf{Accuracy} & \textbf{Drop} & \textbf{Interpretation} \\
\hline
None (Full Set) & 0.9387 & 0.0000 & Baseline model with all selected features. \\
\hline
HSV-LCH & 0.9442 & -0.0056 & Removing improved accuracy — possibly redundant or noisy. Excluded in final set. \\
\hline
RGB-CCM & 0.9399 & -0.0012 & Slight gain in accuracy — not essential. Excluded. \\
\hline
HSV-LBP & 0.9380 & 0.0006 & Very minor drop — mildly useful. Retained. \\
\hline
LAB-CM & 0.9362 & 0.0025 & Small drop — low-impact contributor. Retained. \\
\hline
LAB-LBP & 0.9275 & 0.0112 & Significant drop — most important feature in this set. Retained.\\
\hline
\end{tabular}
\caption{Ablation study on combined color and texture features}
\end{table}

\subsection{Global Feature Set}
To validate the importance of the selected global feature set (LAB-CM, HSV-LBP, and LAB-LBP), we conducted a final ablation study. The full model achieved an accuracy of 95.60\%. Removing any of the three features resulted in a noticeable drop in performance: removing HSV-LBP caused the highest decrease (2.04\%), followed by LAB-Moments (0.99\%), and LAB-LBP (0.68\%). These results confirm that all selected features make meaningful contributions to the classification model and should be retained for the final feature set. No redundant features were identified at this stage.

\begin{table}[H]
\centering

\begin{tabular}{|l|c|c|p{5cm}|}
\hline
\rowcolor{gray!20}
\textbf{Removed Feature} & \textbf{Accuracy} & \textbf{Drop} & \textbf{Interpretation} \\
\hline
None (Full Set) & 0.9560 & 0.0000 & Baseline accuracy with all features included. \\
\hline
HSV-LBP & 0.9356 & 0.0204 & Largest performance drop — most critical feature. \\
\hline
LAB-Moments & 0.9461 & 0.0099 & Significant impact — contributes meaningfully to performance. \\
\hline
LAB-LBP & 0.9492 & 0.0068 & Slight performance drop — still an important feature. \\
\hline
\end{tabular}
\caption{Final ablation study on Global Feature Set}
\end{table}

To make our classification model complete we need to apply feature selector to reduce dimensionality, eliminate redundancy, and enhance classification accuracy. Techniques such as XGBoost selector, LASSO, Random Forest feature importance, and SelectKBest were employed to retain only the most informative features. This process also helped reduce memory usage and training time, ensuring efficient deployment in resource-constrained environments.

\begin{table}[H]
\centering
\resizebox{\textwidth}{!}{%
\begin{tabular}{|l|lllll|}
\hline
\multirow{2}{*}{Setting Name} &
  \multicolumn{5}{l|}{Classification accuracy with different feature selector} \\ \cline{2-6} 
 &
  \multicolumn{1}{l|}{Without feature selector} &
  \multicolumn{1}{l|}{Xgboost} &
  \multicolumn{1}{l|}{Lasso} &
  \multicolumn{1}{l|}{RF} &
  Skbest \\ \hline
Global Feature Set &
  \multicolumn{1}{l|}{95.60} &
  \multicolumn{1}{l|}{95.48} &
  \multicolumn{1}{l|}{92.26} &
  \multicolumn{1}{l|}{95.85} &
  94.42 \\ \hline
\end{tabular}
}
\caption{Most prioritized settings with different feature selectors}
\label{tab:feature_selector_accuracy}
\end{table}
While classification accuracy remained nearly unchanged before and after applying feature selection (Xgb and RF), we opted to retain the feature-selected version due to its reduced dimensionality, lower computational load, and improved model interpretability. This decision ensures a more efficient and scalable deployment, particularly for memory-constrained environments. Where Xgb and RF both have same performance, hence any of them can be choosen for our model.

\subsection{Comparison of Proposed Model with Deep Learning Model}

To compare our proposed model with the deep-learning model described in \cite{machuve2022poultry}, we considered the time taken to execute the code, peak GPU usage, and consumed compute units in Google Colab.   We used the Google Colab Pro version and ran the code using T4 High RAM GPU mode to calculate the computation unit.
\begin{table}[H]
\centering
\resizebox{\textwidth}{!}{%
\begin{tabular}{|l|c|c|c|c|}
\hline
\textbf{Model Name}   & \textbf{Classification Accuracy} & \textbf{Peak GPU} & \textbf{Time} & \textbf{Computation Unit} \\ \hline
Proposed model        & 95.85\%                           & CPU 5.21\%         & 638s          & 0.27                       \\ \hline
Xception              & 96.45\%                          & 13.9 GB            & 8788s         & 4.37                       \\ \hline
MobileNetV3           & 95.22\%                          & 8.2 GB             & 48082s        & 24.35                      \\ \hline
InceptionV3           & 88.99\%                          & 8.1 GB             & 12358s        & 7.6                        \\ \hline
\end{tabular}%
}
\caption{Model comparison}
\label{tab:model_comparison}
\end{table}

\subsection{Evaluation of Proposed Model} The proposed model with Global features extracts important features from HSV and LAB color channels. After extracting the features, we applied the MinMaxScaler function to normalize them. Then, we used principal component analysis for dimensionality reduction with 66 principal components. This helps extract the most relevant features from the data. Later, implementing the XGBoost feature selector allows the most essential features to be selected, where we choose the top 66 features and for RF feature selector 250 with n\_estimators gives the best classification accuracy. Then, we applied the train\_test\_split function to split the dataset for training and validation purposes. We selected test size 0.2, training size 0.8 and random state 44 to train the model with ANN. We added two dense layers to our neural network. The first dense layer has 256 neurons, followed by a dropout layer that randomly drops 50\% of the neurons during training. The second dense layer has 128 neurons, followed by a dropout layer with a 50\% dropout rate. The ReLU activation function is applied in both dense layers. We applied loss 'sparse\_categorical\_crossentropy' to determine the loss function, 'adam' as an optimizer, and metrics as 'accuracy'. We took the early stopping method in the ANN training, a regularization technique used in training machine learning models, especially in iterative algorithms like neural networks. Early stopping aims to prevent overfitting by halting the training process before the model becomes too fitted to the training data. The epoch for the training was 100, and the batch size was 32.

 After the accomplishment of training, we achieved a classification Accuracy of 95.85\% with Rf feature selector and 95.48\% accuracy for Xgb feature selector. The classification report is as follows, 
 \begin{table}[ht]
    \centering
    \resizebox{\textwidth}{!}{%
    \begin{tabular}{lccccc}
        \toprule
        & Precision & Recall & F1-Score & Support & Interpretation \\
        \midrule
        Cocci    & 0.98 & 0.97 & 0.98 & 502 & Model classifies cocci with very high accuracy\\
        Healthy  & 0.92 & 0.96 & 0.94 & 466 & Model performs well, but some misclassifications occur.\\
        NCD      & 0.91 & 0.85 & 0.88 & 111 & Lower recall (0.87) suggests some NCD cases are misclassified. \\
        Salmo    & 0.97 & 0.96 & 0.96 & 535 & Strong classification, but minor misclassifications exist. \\
        \midrule
        Accuracy &      &      & 0.95 & 1614 & 95\% of all test samples were correctly classified. \\
        Macro Avg & 0.95 & 0.93 & 0.94 & 1614 & Unweighted mean, treating all classes equally.\\
        Weighted Avg & 0.96 & 0.95 & 0.95 & 1614 & Weighted by class support (more important for imbalanced datasets).\\
        \bottomrule
    \end{tabular}%
    }
\caption{ANN Classification Report}
\label{tab: classification report}
\end{table}

This table \ref{tab: classification report} presents the classification performance of the model with ANN in classifying poultry diseases across four classes: Cocci, Healthy, NCD, and Salmo. It provides key evaluation metrics: Precision, Recall, F1-Score, and Support for each class, along with overall performance indicators.

Confusion Matrix as given,

\begin{figure}[H]
     \centering
     \includegraphics[width=.8\linewidth]{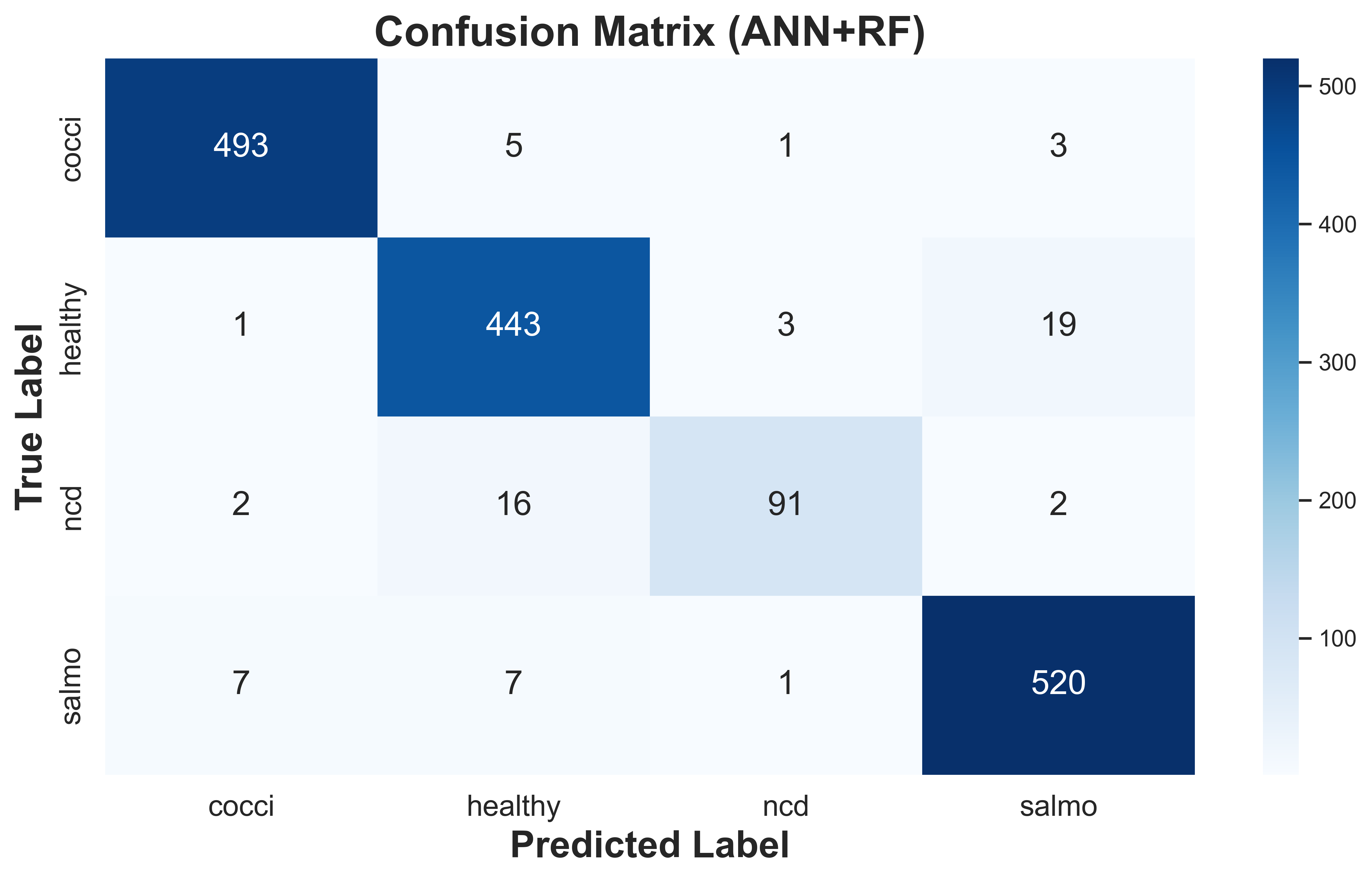}
     \caption{Confusion Matrix }
     \label{fig:confusionmatrix}
 \end{figure}
\textbf{}
Figure \ref{fig:confusionmatrix} is a confusion matrix that visualizes the performance of a classification model in predicting four poultry disease classes: cocci, healthy, ncd, and salmo. The matrix shows that the model performs well, with most predictions aligning with the actual classes (high diagonal values). Where, 484 cocci, 443 healthy, 97 ncd, and 500 salmo images were correctly classified from the test dataset.

 The training accuracy and loss function are,
 \begin{figure}[H]
     \centering
     \includegraphics[width=\linewidth]{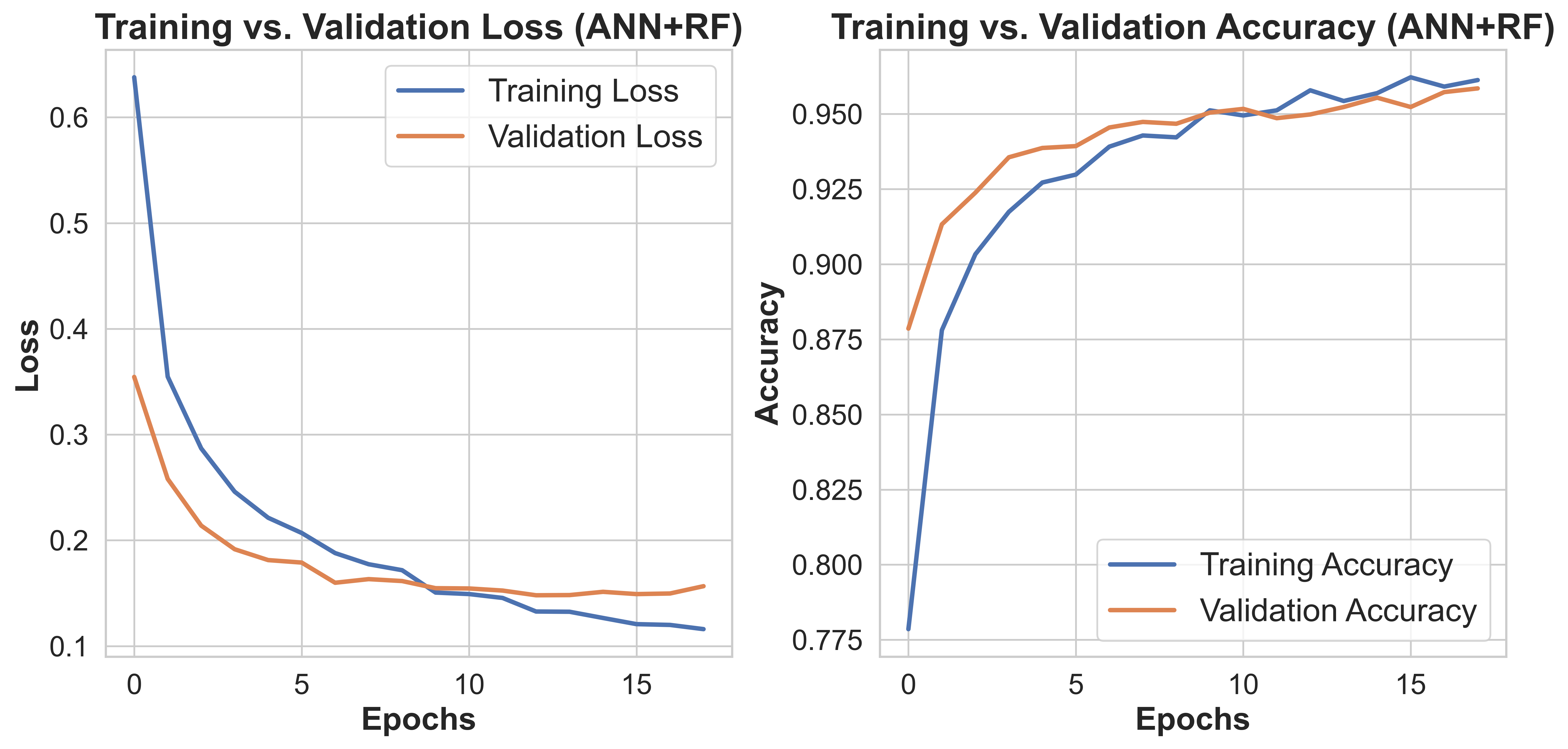}
     \caption{Validation Accuracy and Loss }
     \label{fig:training_loss}
 \end{figure}

Figure \ref{fig:training_loss} consists of two plots comparing training and validation performance over six epochs. The epoch numbers were controlled using the $EarlyStopping$ function with $patience=3$ and $verbose=1$.  The left plot displays Training vs. Validation Loss, where both losses decrease smoothly, indicating the model is learning.  The right plot shows Training vs. Validation Accuracy, where training accuracy steadily increases, while validation accuracy also rises but starts to plateau. This indicates that the model generalizes well up to a point but may struggle with further improvement. Overall, the model performs well.

\subsection{Classification of  selected features with SVM, RF, and DT }

\begin{table}[ht]
  \centering
\begin{tabular}{|l|l|l|l|l|}
\hline
Algorithms & Accuracy & Precision & Recall  & F1-score     \\ \hline
ANN        & 95.85\%  & 95.84\%   & 95.85\%  & 95.81\% \\ \hline
SVM        & 90.71\% & 91.10\%   & 90.71\% & 90.46\% \\ \hline
RF         & 91.82\% & 92.19\%   & 91.82\% & 91.38\% \\ \hline
DT         & 85.44\%  & 85.18\%    & 85.44\%  & 85.24\% \\ \hline
\end{tabular}
\caption{Comparison of ANN with other classification algorithms}
\label{tab:ANN VS others}
\end{table}
Table \ref{tab:ANN VS others} compares the classification performance of the selected global feature set using four different machine learning algorithms: Artificial Neural Network (ANN), Support Vector Machine (SVM), Random Forest (RF), and Decision Tree (DT). Among these, ANN achieves the highest accuracy (95.85\%), precision (95.84\%), recall (95.85\%), and F1-score (95.81\%), demonstrating its superior ability to capture complex patterns in the optimized feature set. In contrast, DT shows the lowest performance, highlighting its limited capacity for multiclass fecal image classification in this context.

\section{Discussion}

The experimental results demonstrate that the proposed lightweight machine learning model achieves high accuracy (95.85\%) in classifying poultry diseases from fecal images while requiring significantly fewer computational resources compared to deep learning models. This performance was achieved by carefully selecting a global feature set based on color (LAB-CM), texture (LAB-LBP and HSV-LBP) descriptors across different color spaces (RGB, HSV, and LAB), optimized via ablation studies and feature selection techniques such as PCA and XGBoost.

Unlike GPU-dependent deep learning methods, our approach is designed for environments with limited hardware capacity, such as rural farms or small-scale poultry operations. The use of interpretable features also makes the system more transparent and adaptable, which is crucial for real-world deployment and further improvement.

The misclassification of certain samples—particularly for the NCD class suggests that class imbalance and visual similarity between disease types may affect precision and recall. Nevertheless, the proposed approach provides a strong balance between accuracy and interpretability, making it suitable for diagnostic applications in resource-constrained settings.

\section{Future Work}

Several promising directions remain for future exploration:

\begin{itemize}
    \item \textbf{Data Augmentation and Balancing:} Future work can focus on addressing class imbalance, especially for underrepresented classes like NCD, using synthetic oversampling or data augmentation techniques.
    
    \item \textbf{Mobile Deployment:} Given the model’s low resource requirements, implementing a real-time mobile or edge-computing application for farmers could extend the impact of this work.
    
    \item \textbf{Multimodal Input:} Integrating other input modalities, such as environmental data (e.g., temperature, humidity) or sound-based analysis, could enhance detection robustness.
    
    \item \textbf{Explainable AI:} Incorporating explainability tools (e.g., SHAP or LIME) would help users understand model predictions and improve trust among practitioners.
    
    \item \textbf{Extended Disease Classes:} As more annotated datasets become available, expanding the system to classify a wider range of poultry diseases will enhance its applicability and usefulness.
    
    \item \textbf{Cross-Farm Validation:} Validating the model across datasets collected from different geographic regions or farm conditions would further demonstrate generalizability.
\end{itemize}





\section*{Declaration of Generative AI and AI-assisted Technologies in the Writing Process}
During the preparation of this work, the authors used generative AI tools (ChatGPT, DeepSeek, Grammarly) to assist with language refinement, grammatical correction, formatting suggestions, and manuscript editing. After using this tool, the authors reviewed and edited the content as necessary and take full responsibility for the integrity and accuracy of the content of this manuscript.


\end{document}